\documentclass[pdflatex,sn-mathphys-num,referee]{sn-jnl}% Math and Physical Sciences Numbered Reference 
\usepackage{mathrsfs}
\usepackage{graphicx}%
\usepackage{multirow}%
\usepackage{amsmath,amssymb,amsfonts}%
\usepackage{amsthm}%
\usepackage[title]{appendix}%
\usepackage{xcolor}%
\usepackage{textcomp}%
\usepackage{manyfoot}%
\usepackage{booktabs}%
\usepackage{algorithm}%
\usepackage{algorithmicx}%
\usepackage{algpseudocode}%
\usepackage{listings}%
\usepackage{blindtext}%
\usepackage{tabularx}%
\usepackage{array}%
\usepackage{tikz}%
\usetikzlibrary{shapes.geometric, arrows}%
\usepackage{algorithm}
\usepackage{setspace} % for \setstretch
\usetikzlibrary{shapes.geometric, arrows.meta, positioning}

\usepackage{float}  % Provides H float placement
\usepackage[labelsep=space]{caption} % Essential for \ContinuedFloat
\usepackage{afterpage}  % For \afterpage
\usepackage{placeins}  % For \FloatBarrier
\usepackage{booktabs}
\usepackage{orcidlink}

\theoremstyle{thmstyleone}%
\theoremstyle{thmstyletwo}%

\theoremstyle{thmstylethree}%
\usepackage{algorithm}
\usepackage{algpseudocode}  % Corrected algorithmic environment
\usepackage{amsmath}
\usepackage{amssymb}

\tikzstyle{start stop} = [rectangle, rounded corners, minimum width=2.5 cm, minimum height=1 cm, text centered, draw=black, fill=red!30]
\tikzstyle{Io} = [trapezium, trapezium left angle=70, trapezium right angle=110, minimum width=2 cm, minimum height=1 cm, text width=3 cm, text centered, draw=black, fill=blue!30]
\tikzstyle{process} = [rectangle, minimum width=1 cm, minimum height=1 cm, draw=black, text width=6 cm, fill=orange!30]
\tikzstyle{decision} = [diamond, minimum width=3 cm, minimum height=1 cm, text centered, draw=black, fill=green!30]
\tikzstyle{arrow} = [thick, ->, >=stealth]

\begin{document}

\title{Diversity Conscious Refined Random Forest}

\author[1]{Sijan Bhattarai\textsuperscript{*}}\email{sijan.762419@thc.tu.edu.np}
\author[1]{Saurav Bhandari\textsuperscript{*}}\email{saurav.762419@thc.tu.edu.np}
\author[2]{Girija Bhusal\textsuperscript{\dag}}\email{girija.bhusal9@gmail.com}
\author[1]{Saroj Shakya\textsuperscript{\dag}}\email{sarojsh@tcioe.edu.np}
\author[3]{Tapendra Pandey\textsuperscript{\dag}}\email{pandey@ou.edu}

\affil[1]{Department of Electronics and Computer Engineering, Institute of Engineering, Thapathali Campus, Tribhuvan University, Kathmandu, Nepal}
\affil[2]{Department of Computer Science, Swastik College, Tribhuvan University, Bhaktapur, Nepal}
\affil[3]{School of Computer Science, Gallogly College of Engineering, University of Oklahoma, Norman, USA}


\abstract{Random Forest (RF) is a widely used ensemble learning technique known for its robust classification performance across diverse domains. However, it often relies on hundreds of trees and all input features, leading to high inference cost and model redundancy. In this work, our goal is to grow trees dynamically only on informative features and then enforce maximal diversity by clustering and retaining uncorrelated trees. Therefore, we propose a Refined Random Forest Classifier that iteratively refines itself by first removing the least informative features and then analytically determines how many new trees should be grown, followed by correlation-based clustering to remove redundant trees. The classification accuracy of our model was compared against the standard RF on the same number of trees. Experiments on 8 multiple benchmark datasets, including binary and multiclass datasets demonstrate that the proposed model achieves improved accuracy compared to standard RF.}

%%================================%%
%% Sample for structured abstract %%
%%================================%%

\keywords{Random Forest,Ensemble, Refined Random Forest, Correlation, Pruning}

\maketitle

% Place this after \maketitle
\let\thefootnote\relax
\footnotetext{\textsuperscript{*}These authors contributed equally to this work.}
\footnotetext{\textsuperscript{\dag}Corresponding authors: girija.bhusal9@gmail.com, sarojsh@tcioe.edu.np, pandey@ou.edu}

\section{Introduction} \label{sec1}

RF is a popular ensemble learning algorithm known for its simplicity, strong predictive performance, and generalization to diverse tasks \cite{bib101}. Its robustness to overfitting, the ability to handle high-dimensional data and the estimation of inherent characteristics importance have made it popular in many domains such as QSAR modeling in cheminformatics \cite{bib102}, land cover and hyperspectral image classification in remote sensing \cite{bib103}, and merging of multiple satellite precipitation products in hydrology \cite{bib104}. However, RF often uses hundreds of trees, leading to high training and inference latency, high memory usage, and redundancy when many trees convey similar information.

This study investigates the use of RFs through a quantitative approach, evaluating the model on  binary and multiclass datasets, a method that dynamically adjusts the number of trees to identify the optimal forest size was proposed in \cite{bib111}. Similarly, a technique for selecting the most diverse trees within a RF to reduce redundancy was introduced in \cite{bib113}. Building on these ideas, our study explores whether dynamically adjusted forests still contain correlated trees that produce nearly identical probability distributions across class labels, indicating redundancy. We aim to develop an algorithm that not only adjusts the number of trees dynamically based on the dataset but also removes these correlated trees. The resulting model is called the Diversity-Conscious Refined Random Forest (DCRRF).

Experiments have shown that the error rate may not decrease monotonically with the addition of more trees and can sometimes even increase, suggesting that more trees are not always better \cite{bib105}. Hyperparameter tuning strategies have been proposed to emphasize the importance of selecting an appropriate number of trees, among other parameters, to improve model performance \cite{bib106}. Several other studies have aimed to make RFs more robust and resource efficient. For example, large ensembles are not always necessary to achieve comparable levels of consistency \cite{bib107}. Furthermore, shallower trees can act as a form of regularization, which could reduce the need for deeper and larger forests \cite{bib108}.

An analysis of optimal settings for classification tree ensembles in medical decision support suggests that using a large number of shallow trees can yield better results \cite{bib101}. This highlights the importance of optimizing both the tree size and the number of trees. Additionally,the Random Forest-based method to merge satellite, reanalysis, and topographic data with ground rain gauge measurements to improve precipitation estimates. Applied in Chile (2000–2016), RF-MEP outperformed existing datasets and merging techniques, even with limited training data\cite{bib109}.

The paper "Improved Random Forest for Classification" introduces a variant of Random Forest that minimizes ensemble size by iteratively separating features into “important” and “unimportant” sets based on global feature weights, and pruning the latter in each {iteration}. They then derive a theoretical upper bound on the number of new trees to add—based on the counts of important vs. unimportant features to guarantee a net accuracy gain, and prove that once this bound is reached, further tree growth or feature pruning no longer improves performance \cite{bib111}. We adopt the technique of separating features into “important” and “unimportant” groups, along with feature pruning and tree addition strategies. This approach has also been successfully applied to text classification, where the same pruning and growth strategy was used to reduce feature dimensionality and improve performance across multiple corpora ~\cite{bib112}.

Separately, tree redundancy has been addressed by clustering trees based on their pairwise output correlations and selecting the highest performing tree from each cluster using permutation-based importance~\cite{bib113}. Inspired by this, we build correlation clusters in our interim forests, compute each tree’s AUC using a separate part of the data for validation, and select only the highest AUC tree per cluster, producing a compact and maximally diverse final ensemble. Similarity,  other de-correlation strategies include Extremely Randomized Trees which randomize split thresholds to lower inter-tree \cite{bib114}, and Adaptive Random Forests which dynamically adjust tree counts and feature subsets in streaming data contexts \cite{bib115}.

We propose a Diversity Concious Self-Growing Random Forest called the Refined Random Forest (RRF). RRF dynamically adapts the model by iteratively pruning less informative features and growing new trees only when necessary, based on the current model's performance. Furthermore, it introduces a correlation-based clustering mechanism to retain only the most diverse and uncorrelated trees in the final ensemble. This ensures both efficiency and improved generalization performance. Our method incorporates concepts from previous research on adaptive forests, such as feature refinement, iterative tree addition, and correlation-based pruning and applies them in a our framework \cite{bib111, bib113}. We evaluated our approach on eight benchmark datasets, covering both binary and multiclass classification tasks, and demonstrated consistent performance improvements over the standard Random Forest, using the same number of trees.

The remainder of the paper is organized as follows. In Section II we describe the proposed methodology in detail. The experimental details,
results and related discussions are presented in Section III and finally we conclude the paper in Section IV.

\section{Method} \label{sec2}
% In our study, we opt to choose information gain rather than quality of split. This is because………….. 
The proposed Refined Random Forest builds on previous work by adaptively growing the forest through iterative addition of trees and systematic feature selection. In each iteration, feature importance is evaluated to remove consistently low-weight features while retaining those deemed important. A controlled upper limit on new trees ensures continuous performance improvement, and the process stops when gains become minimal. This method includes a final refinement step that eliminates correlated trees to improve ensemble diversity, resulting in a more robust, efficient, and better-generalizing model.The steps are given below in detail.

The enhanced version incorporates an additional refinement step after the final stage to obviate correlated trees, extending the Improved Random Forest approach introduced in \cite{bib111}. These trees contribute little to ensemble diversity and are therefore filtered out. By retaining only the uncorrelated, diverse trees that offer unique contributions to the final  decision, the Refined Random Forest achieves better generalization and efficiency, resulting in a more robust and optimized model.
% ---------------------------------------------------------start symbols--------------------------------
\FloatBarrier
\begin{table}[h]
\caption{Symbol definitions used in the proposed method}\label{tab:symbols}
\renewcommand{\arraystretch}{1.5}
\begin{tabular*}{\textwidth}{@{\extracolsep\fill}ll}
\toprule
\textbf{Symbol} & \textbf{Definition} \\
\midrule
$F_0$ & Initial Feature Vector \\
$F_n$ & Feature vector after $n^{\text{th}}$ iteration \\
$\gamma_n$ & Forest after $n^{\text{th}}$ iteration \\
$\tau_n$ & Trees after $n^{\text{th}}$ iteration \\
$I$ & Bag of important features \\
$U$ & Bag of unimportant features \\
% $\omega(t)$ & Weight of feature $t$ \\
$R$ & Features removed after $n^{\text{th}}$ iteration \\
$A$ & Features added after $n^{\text{th}}$ iteration \\
$\lambda$ & Set of features selected for node split \\
$f$ & Number of selected features for node split \\
$\rho$ & Probability of selecting an important feature \\
$q$ & Probability of selecting no important feature \\
$\zeta$ & Strength of forest \\
$T_{av}$ & Average number of nodes per tree \\
$\prod$ & Probability that at least one feature is common among corresponding nodes of two trees \\
$\chi$ & Classification Accuracy \\
$C$ & Correlation among trees \\
$w^{\tau}(j)$ & Local weight of feature $j$ in tree $\tau$ \\
$\Delta u$ & Change in number of important features \\
$\Delta v$ & Change in number of unimportant features \\
$\Delta B$ & Number of trees to add at iteration $n$ \\
$\delta^{\tau}$ & Out‑of‑bag error of tree $\tau$ \\

\bottomrule
\end{tabular*}
\end{table}
\FloatBarrier
%-----------------------------------------end symbols---------------------------------------------

%-----------------------------------------start algorithm------------------------------------------

\FloatBarrier
\begin{algorithm}
\caption{Refined Random Forest Algorithm}
\begin{algorithmic}[1]

\State Initialize the initial forest $\gamma_{0}$ with $T_{0}$ trees using feature vector $F_{0}$.
\State Compute global weights $\omega(t)$ of each from $F_{0}$ using Equation (1).
\State Rank features based on $\omega(t)$ using Equation (2):
\begin{itemize}
    \item Select the top $\lfloor \sqrt{|F_{0}|} \rfloor$ features as important features, and assign them to $I_{0}$.
\end{itemize}

\State Assign remaining features to $U_{0}$ (unimportant feature bag).
\State Assign iteration counter $n = 0$.
\State Compute mean ($\mu_{n}$) and standard deviation ($\sigma_{n}$) of feature weights $\{\omega(t):\,t \in U_{n}\}$.
\State Identify features to remove from $U_{n}$:
\[
  R \;=\;
    \begin{cases}
      \{\,t \in U_{n} : \omega(t) < \mu_{n} - 2\,\sigma_{n}\}, & \text{if nonempty},\\
      \{\,t \in U_{n} : \omega(t) = \min_{k \in U_{n}} \omega(k)\}, & \text{otherwise}.
    \end{cases}
\]
\State Identify features to promote $A$ from $U_{n}$ to $I_{n}$:
\[
  A \;=\; \{\,t \in U_{n} : \omega(t) \ge \min_{k \in I_{n}} \omega(k)\}.
\]
\State Update feature set and bags: 
\[
  F_{\,n+1} = F_{n} \setminus R,\quad
  I_{n+1} = I_{n} \cup A,\quad
  U_{\,n+1} = U_{\,n} \setminus (A \cup R).
\]
\State Compute changes: $\Delta I = |I_{n+1}| - |I_{n}|,\; \Delta U = |U_{\,n+1}| - |U_{\,n}|$.
\State Compute $\Delta \tau$ using Equation (4), then update $\tau_{\,n+1} = \tau_{\,n} + \Delta \tau$.
\State Grow new forest $\gamma_{\,n+1}$ with $\tau_{\,n+1}$ trees using feature vector $F_{\,n+1}$.
\State Compute updated $\omega(t)$ for all features and re‐rank using Equation (2).
\State $n \leftarrow n + 1$.
\State Use each tree to independently predict probabilities (or class labels) on the test set:
\begin{itemize}
    \item Collect all individual tree predictions: $\hat{p}_{1}, \hat{p}_{2}, \ldots, \hat{p}_{T}$
\end{itemize}
\State Compute AUC per tree:
\begin{itemize}
    \item For each tree $t \in \{1, \ldots, T\}$, compute the AUC score $\mathrm{AUC}_{t} = \mathrm{AUC}(\hat{p}_{t},\,y_{\mathrm{true}})$.
\end{itemize}
\State Compute the pairwise correlation matrix among prediction vectors:
\[
    \rho_{\,i,j} = \mathrm{Corr}(\hat{p}_{i},\,\hat{p}_{j}) \quad \forall\,i, j \in \{1, \ldots, T\}.
\]
\State Apply correlation-based clustering:
\begin{itemize}
    \item Group trees such that trees within a cluster are highly correlated.
    \item Ensure clusters are as uncorrelated as possible using a threshold $\text{th} \ge 0.93$.
\end{itemize}
\State From each cluster, select the tree with the highest AUC score.
\State Use the selected $K$ trees to perform final ensemble prediction.

\end{algorithmic}
\end{algorithm}
\FloatBarrier
% %-----------------------------------------------end algorithm-------------------------------------

%---------------------------------start flowchart------------------------------

\subsection{Flowchart} 
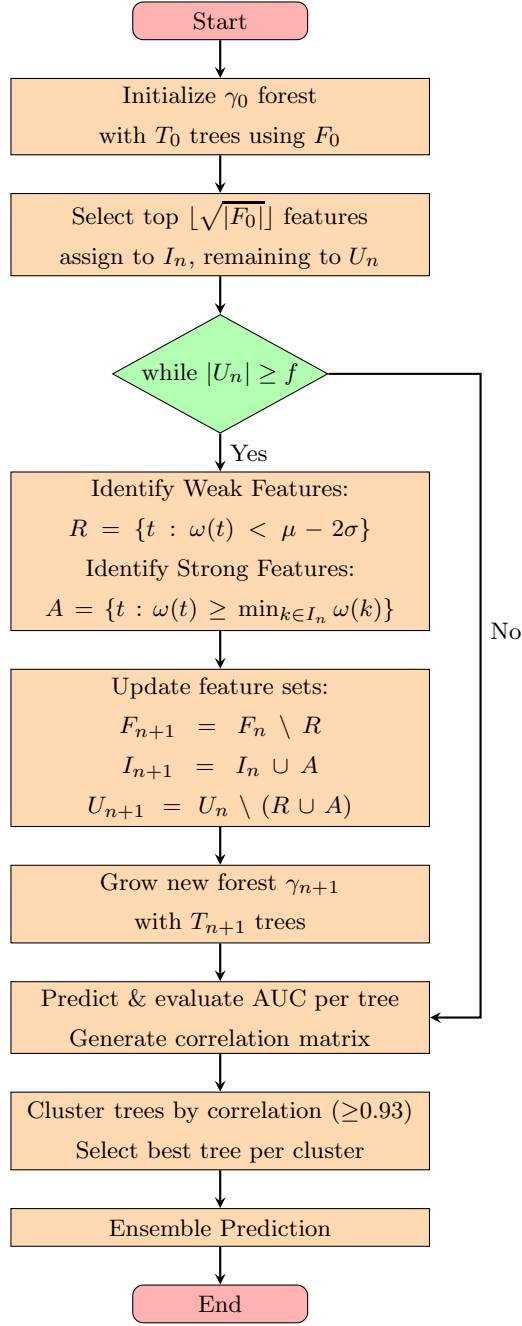
\begin{figure}[H]
\centering
\begin{tikzpicture}[
    node distance=0.5 cm,  % Reduced from 0.65cm
    every node/.style={font=\scriptsize},  % Smaller font
    start stop/.style={rectangle, rounded corners, draw, fill=red!30, minimum width=2.3 cm, minimum height=0.5 cm, align=center},
    process/.style={rectangle, draw, fill=orange!30, text width=5.3 cm, minimum height=0.5 cm, align=center, inner sep = 2.9 pt},  % Tighter padding
    decision/.style={diamond, draw, fill=green!30, aspect=1.8, minimum width=1.6 cm, minimum height=0.8 cm, align=center, inner sep=1 pt},
    arrow/.style={thick,->,>=stealth}
]

% Nodes with compact formatting
\node (start) [start stop] {Start};

\node (initialize_forest) [process, below=of start] {
    Initialize $\gamma_0$ forest \\ with $T_0$ trees using $F_0$
};

\node (select_feature) [process, below=of initialize_forest] {
    Select top $\lfloor\sqrt{|F_0|}\rfloor$ features \\ assign to $I_n$, remaining to $U_n$
};

\node (decision) [decision, below=of select_feature] {while $|U_n| \geq f$};

\node (if_yes) [process, below=of decision] {
    Identify Weak Features: \\
    $R = \{t:\omega(t) < \mu - 2\sigma\}$ \\
    Identify Strong Features: \\
    $A = \{t:\omega(t) \geq \min_{k\in I_n}\omega(k)\}$
};

\node (imp_unimp) [process, below=of if_yes] { 
    Update feature sets: \\
    $F_{n+1} = F_n \setminus R$ \\
    $I_{n+1} = I_n \cup A$ \\
    $U_{n+1} = U_n \setminus (R\cup A)$
};

\node (grow_new_forest) [process, below=of imp_unimp] {
    Grow new forest $\gamma_{n+1}$ \\ with $T_{n+1}$ trees
};

\node (evaluate_auc) [process, below=of grow_new_forest] {
    Predict \& evaluate AUC per tree \\ Generate correlation matrix
};

\node (form_cluster) [process, below=of evaluate_auc] {
    Cluster trees by correlation ($\geq$0.93) \\ Select best tree per cluster
};

\node (ensemble) [process, below=of form_cluster] {Ensemble Prediction};

\node (end) [start stop, below=of ensemble] {End};

% Connections
\draw [arrow] (start) -- (initialize_forest);
\draw [arrow] (initialize_forest) -- (select_feature);
\draw [arrow] (select_feature) -- (decision);
\draw [arrow] (decision) -- node[right] {Yes} (if_yes);
\draw [arrow] (if_yes) -- (imp_unimp);
\draw [arrow] (imp_unimp) -- (grow_new_forest);
\draw [arrow] (grow_new_forest) -- (evaluate_auc);
\draw [arrow] (evaluate_auc) -- (form_cluster);
\draw [arrow] (form_cluster) -- (ensemble);
\draw [arrow] (ensemble) -- (end);

% "No" branch
\draw [arrow] (decision.east) -- ++(2,0) |- node[pos=0.2, above,anchor= west] {No} (evaluate_auc);

% \draw [arrow] (decision.east) -- ++(2.5,0) |- (evaluate_and_cluster.east) node[near start, anchor=west] {no};

\end{tikzpicture}
\caption{Compact flowchart of the feature-selection process}
\label{fig:flowchart}
\end{figure}

%---------------------------------------------end flowchart---------------------------------------
\subsection{Feature Ranking}

At iteration \(n\), let the current feature set be \(F_{n}\) and let the forest \(\gamma_{n}\) consist of \(B_{n}\) trees. Our goal is to assign each feature \(j \in F_{n}\) a global importance score \(w(j)\in [0,1]\) that reflects how discriminative that feature is across all trees. We obtain \(w(j)\) by combining three components: the local split‐quality weight of feature \(j\) in each tree \(\tau\), a normalization factor based on each tree’s out‐of‐bag error \(\delta^{\tau}\), and a final aggregation and normalization across all features.

\subsubsection{Local Weight of Feature \(j\) in Tree \(\tau\)}
For a given tree \(\tau\), each internal node \(i\) that splits on feature \(j\) contributes an information‐gain ratio:
\begin{equation}
\mathrm{IGR}(i, j) =
\frac{H(N_{i}) - \dfrac{N_{\ell}}{N_{i}}\,H(N_{\ell}) - \dfrac{N_{r}}{N_{i}}\,H(N_{r})}
     {H(N_{i})}
\end{equation}

where \(H(\cdot)\) denotes Shannon entropy, \(N_{i}\) is the sample count at the parent node \(i\), and \(N_{\ell}, N_{r}\) are the sample counts at the left and right child nodes, respectively.  

Let \(T_{av}\) be the average number of internal nodes per tree (as per the symbol table), and let \(N\) be the total number of internal nodes in tree \(\tau\). Averaging these ratios over all splits in \(\tau\) yields the \emph{local weight} of feature \(j\):
\begin{equation}
w^{\tau}(j) =
\frac{1}{N} \sum_{i=1}^{N} \mathrm{IGR}(i, j)
\end{equation}

A higher value of \(w^{\tau}(j)\) indicates that feature \(j\) consistently provides high‑quality splits throughout tree \(\tau\).\\[0.3em]
\noindent\textbf{Why Information Gain and Gain Ratio over Quality Of Split?}

Information Gain (IG) is a principled metric for node splitting in decision trees, rooted in information theory. It directly quantifies the reduction in entropy after a split:

\begin{equation}
\mathrm{IG}(i, j) = H(N_i) - \left( \frac{N_\ell}{N_i} H(N_\ell) + \frac{N_r}{N_i} H(N_r) \right),
\end{equation}

where \( H(N_i) \) is the entropy of the parent node, and \( H(N_\ell), H(N_r) \) are the entropies of the left and right child nodes, respectively \cite{bib118}. 

In contrast, some methods use a raw ``quality-of-split'' score:

\begin{equation}
Q(i, j) = \exp[-(H(N_\ell) + H(N_r))],
\end{equation}

as seen in Improved Random Forest (IRF) implementations \cite{bib111}. However, \( Q(i,j) \) ignores how impure the parent node was, and treats all nodes equally regardless of how much uncertainty they originally contained. This makes it difficult to distinguish splits that actually reduce uncertainty from those that merely rearrange entropy \cite{bib117}.

Information Gain resolves this by anchoring each split to the parent node’s uncertainty, producing more meaningful splits and trees that generalize better \cite{bib116}.\\[0.3em]

\noindent\textbf{Why Information Gain Ratio(IGR) over Information Gain(IG)?}

Raw Information Gain (IG) is known to be biased toward high-cardinality features. Features with many distinct values can create overly pure splits simply by fragmenting data evenly—even when those splits have no predictive value.

To mitigate this, we use the Information Gain Ratio (IGR):

\begin{equation}
\mathrm{IGR}(i, j) = \frac{\mathrm{IG}(i, j)}{H(N_i)}.
\end{equation}

This normalizes IG by the parent’s entropy \(H(N_i)\), producing a score that reflects the proportion of uncertainty removed by the split. IGR penalizes deceptive splits on high-cardinality features and yields more robust, generalizable models \cite{bib117, bib110}.\\[0.3em]

\noindent\textbf{Why We Avoid Normalizing IG by the Maximum IG in the Forest?}

A naïve alternative is to normalize raw IG by the maximum IG observed anywhere in the forest:

\begin{equation}
\tilde{\text{IG}}(i, j) = \frac{\text{IG}(i, j)}{\max_t \text{IG}(t)}.
\end{equation}

This rescales scores to [0,1], but it creates two key problems:

\textbf{Loss of Interpretability:} This approach no longer tells us how much uncertainty was reduced at node \( i \). In contrast, \( \text{IGR}(i,j) = 0.5 \) implies “50\% of entropy at node \( i \) was eliminated.”

\textbf{Inconsistent Scaling:} Consider a nearly pure node (\( E_i = 0.1 \)) and a highly impure node (\( E_i = 0.9 \)). A 0.05 IG on the pure node may be highly significant but looks insignificant (0.0625) after normalization. Meanwhile, a moderate IG of 0.4 from the impure node appears stronger (0.5), even if less meaningful.

Normalizing by the parent’s entropy preserves proportional meaning and ensures splits are assessed fairly across varying node purities. Thus, we adopt IGR as the more theoretically grounded and empirically consistent approach.

\subsubsection{Tree‑Level Normalization (OOB Weighting)}

Let \(\delta^{\tau}\) be the out‑of‑bag error of tree \(\tau\). Define its normalized weight:
\begin{equation}
\gamma^{\tau} =
\frac{\displaystyle \frac{1}{\delta^{\tau}}}
     {\displaystyle \max_{\tau} \left( \frac{1}{\delta^{\tau}} \right)}
\end{equation}
A smaller \(\delta^{\tau}\) (i.e.\ better OOB accuracy) yields a larger \(\gamma^{\tau}\), so that trees with lower classification error contribute more to the global feature score.

\subsubsection{Global Weight of Feature \(j\)}  
Aggregate the local weights \(w^{\tau}(j)\) across all \(B_n\) trees, weighted by \(\gamma^{\tau}\), and normalize over all features. First define
\begin{equation}
S(j) = \sum_{\tau=1}^{B_n} w^{\tau}(j)\,\gamma^{\tau}
\end{equation}

Then the global importance score is
\begin{equation}
w(j) = 
\frac{\,S(j)\,}
     {\,\max_{\,k \in F_{n}}\!S(k)\,}
\end{equation}
For each \(j \in F_{n}\), features with higher \(w(j)\) are deemed more important for classification.

\subsection{Finding Important and Unimportant Features}

Once every feature \(j \in F_{n}\) has been assigned a global weight \(w(j)\), we partition the current feature set \(F_{n}\) into two pools:
\[
I_n = \{\text{“important” features, of size } u_n \}
\]
\[
U_n = \{\text{“unimportant” features, of size } v_n \}
\]

such that \(u_n + v_n = |F_n|\). Initially (\(n=0\)), $I_0$ consists of the top \(\bigl\lfloor \sqrt{|F_0|} \bigr\rfloor\) features by weight, and \(U_0 = F_0 \setminus I_{n}\). At each subsequent pass \(n\), we refine these pools as follows:

\subsubsection{Candidate Prune Set}
Features whose weights fall more than two standard deviations below the mean are candidates for pruning:
\begin{equation}
R_{n} 
= 
\{ j \in U_{n} : w(j) < \mu_{n} - 2\sigma_{n} \}
\end{equation}

If no such features exist (\( R_{n} = \varnothing \)), then prune the single least important feature in \( U_{n} \):
\begin{equation}
R_{n} 
= 
\{ j \in U_{n} : w(j) = \min_{k \in U_{n}} w(k) \}
\end{equation}

\subsubsection{Promotion of Near‑Threshold Features}
Let the minimum global weight among important features be:
\[
m_{n}
\;=\;
\min_{k \in I_{n}} w(k)
\]

Promote any feature from \(U_{n}\) whose weight is at least \(m_{n}\):
\begin{equation}
A_{n}
= 
\{ j \in U_{n} : w(j) \ge m_{n} \}
\end{equation}

\subsubsection{Update Feature Pools}

Update the feature set by removing pruned features:
\begin{equation}
F_{n+1}
= 
F_{n} \setminus R_{n}
\end{equation}

Update the important feature pool:
\begin{equation}
I_{n+1}
= 
I_{n} \cup A_{n}
\end{equation}

Update the unimportant feature pool:
\begin{equation}
U_{n+1}
= 
U_{n} \setminus (R_{n} \cup A_{n})
\end{equation}

Once a feature is promoted to $I$, it is never removed in future iterations.

\subsubsection{Record Pool Changes}
Track the change in size of the important feature pool:
\begin{equation}
\Delta u = \lvert I_{n+1} \rvert - \lvert I_{n} \rvert
\end{equation}
Track the change in size of the unimportant feature pool:
\begin{equation}
\Delta v = \lvert U_{n+1} \rvert - \lvert U_{n} \rvert
\end{equation}

\subsection{Finding the Number of Trees to Be Added}

After updating the feature pools, we compute the number of new trees \(\Delta B\) to grow such that the overall classification accuracy \(\chi\) strictly increases. This process relies on the concepts of strength \(\zeta\) and correlation \(\eta_c\), as introduced by Breiman \cite{bib101} and extended in IRF \cite{bib111}.

\subsubsection{Probability of a Good Split}

A node split is considered “good” if the randomly sampled set \(\lambda\) of \(f\) features contains at least one important feature. Let \(q\) denote the minimum probability of such a good split. Then

\begin{equation}
q = 
\begin{cases}
1 - \frac{\binom{v_n}{f}}{\binom{u_n + v_n}{f}}, & \text{if } v_n \geq f \text{ and } u_n + v_n \geq f \\
1, & \text{otherwise}
\end{cases}
\end{equation}

\subsubsection{Partial Derivatives of Good Split Probability}

To quantify the effect of feature updates, we compute discrete approximations of the partial derivatives of \(q\):

\begin{equation}
q_u \approx -\frac{\Delta r}{\Delta u} = \frac{v_n! \, (u_n + v_n - 1 - f)! \cdot f}{(v_n - f)! \, (u_n + v_n)!}
\end{equation}
\begin{equation}
q_v \approx -\frac{\Delta r}{\Delta v} = -\frac{(v_n - 1)! \, (u_n + v_n - 1 - f)! \cdot u_n f}{(v_n - f)! \, (u_n + v_n - 1)! \, (u_n + v_n)}
\end{equation}

\subsubsection{Strength and Correlation}

Assuming each tree has \(T_{av}\) nodes, the forest strength is defined as
\begin{equation}
\zeta = 1 - (1 - q^{T_{av}})^B
\end{equation}
and the probability \(p\) that any two trees select a common node split is
\begin{equation}
p = 1 - \frac{\binom{u_n + v_n - f}{f}}{\binom{u_n + v_n}{f}}
\end{equation}
The average correlation between any two trees is
\begin{equation}
C = p^{T_{av}}
\end{equation}
and the correlation component of accuracy is
\begin{equation}
\eta_c = 1 - (1 - C)^{B/2}
\end{equation}

\subsubsection{Classification Accuracy}

Classification accuracy is then expressed as
\begin{equation}
\chi = \lambda (\zeta - \eta_c)
\end{equation}
where \(\lambda\) is a scaling constant.

\subsubsection{Change in Accuracy and Deriving $\Delta B$}

Differentiating $\chi$ and expressing change in accuracy $d\chi$ as
\begin{equation}
d\chi \approx \lambda \left( \nu\, \Delta B + l\, q_u\, \Delta u + l\, q_v\, \Delta v \right)
\end{equation}
where
\[
l = B \cdot \text{Nav} \cdot q^{\text{Nav} - 1} (1 - q^{\text{Nav}})^{B - 1}
\]
and
\[
\nu = \frac{\partial (\zeta - \eta_c)}{\partial B}
\]

To ensure that accuracy improves, we require
\begin{equation}
|\Delta B| \;<\; \left|\frac{\,l\,\bigl(q_u\,\Delta u \;+\; q_v\,\Delta v\bigr)\,}{\nu}\right|.
\end{equation}

Consequently, $\Delta B$ is chosen to satisfy this inequality, with the additional constraint
\[
\Delta B \;\ge\; 0,
\]
thereby guaranteeing that the ensemble’s classification performance does not degrade.

\subsection{Correlation-Based Pruning}

Due to the inherent randomization in Random Forests, trees can become highly correlated, yielding redundant predictions and reducing ensemble diversity~\cite{bib113}. We illustrate how pruning removes these redundancies on the Breast Cancer dataset:

\begin{figure}[h]
    \centering
    \includegraphics[width=0.48\columnwidth]{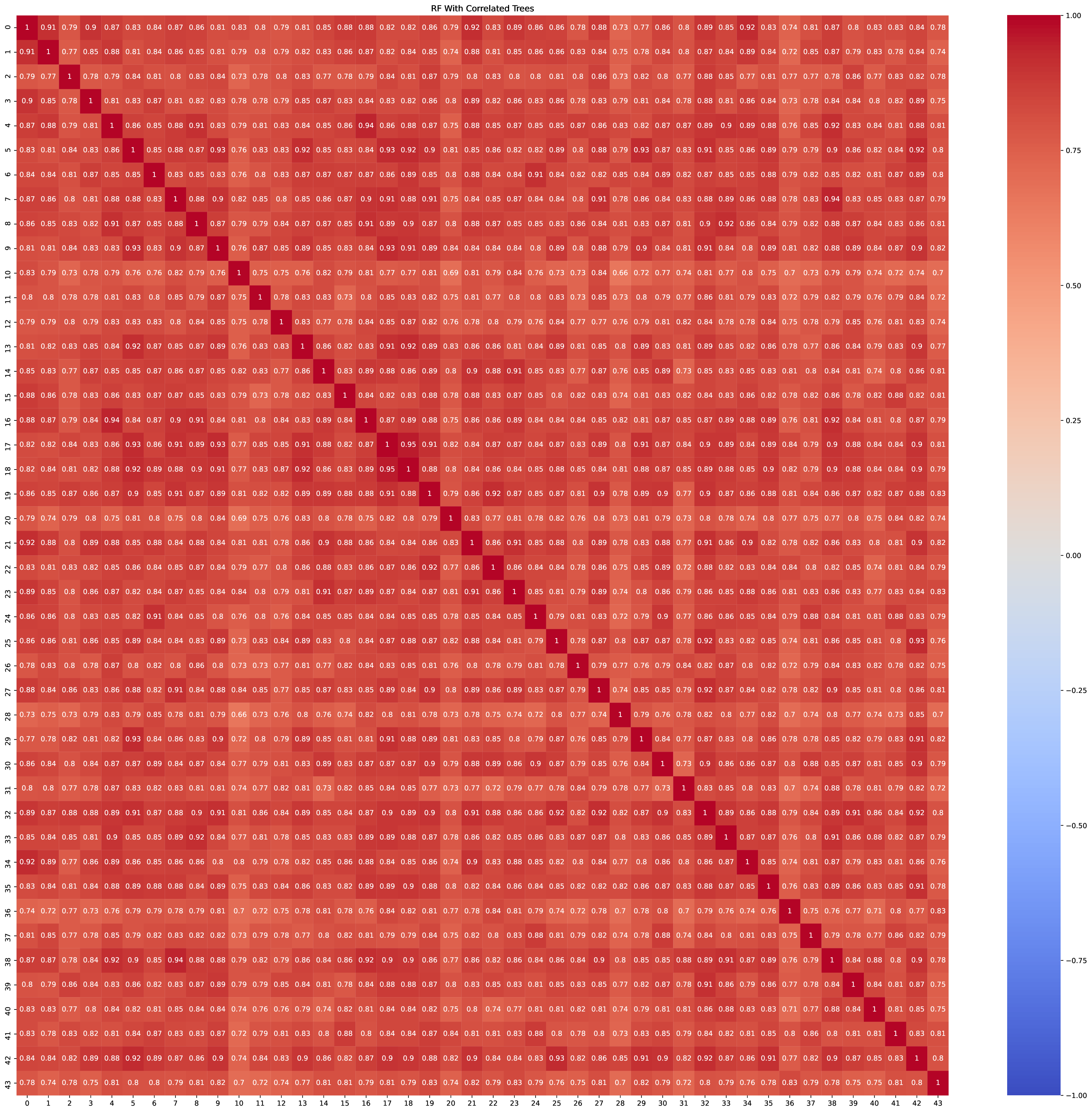}
    \includegraphics[width=0.48\columnwidth]{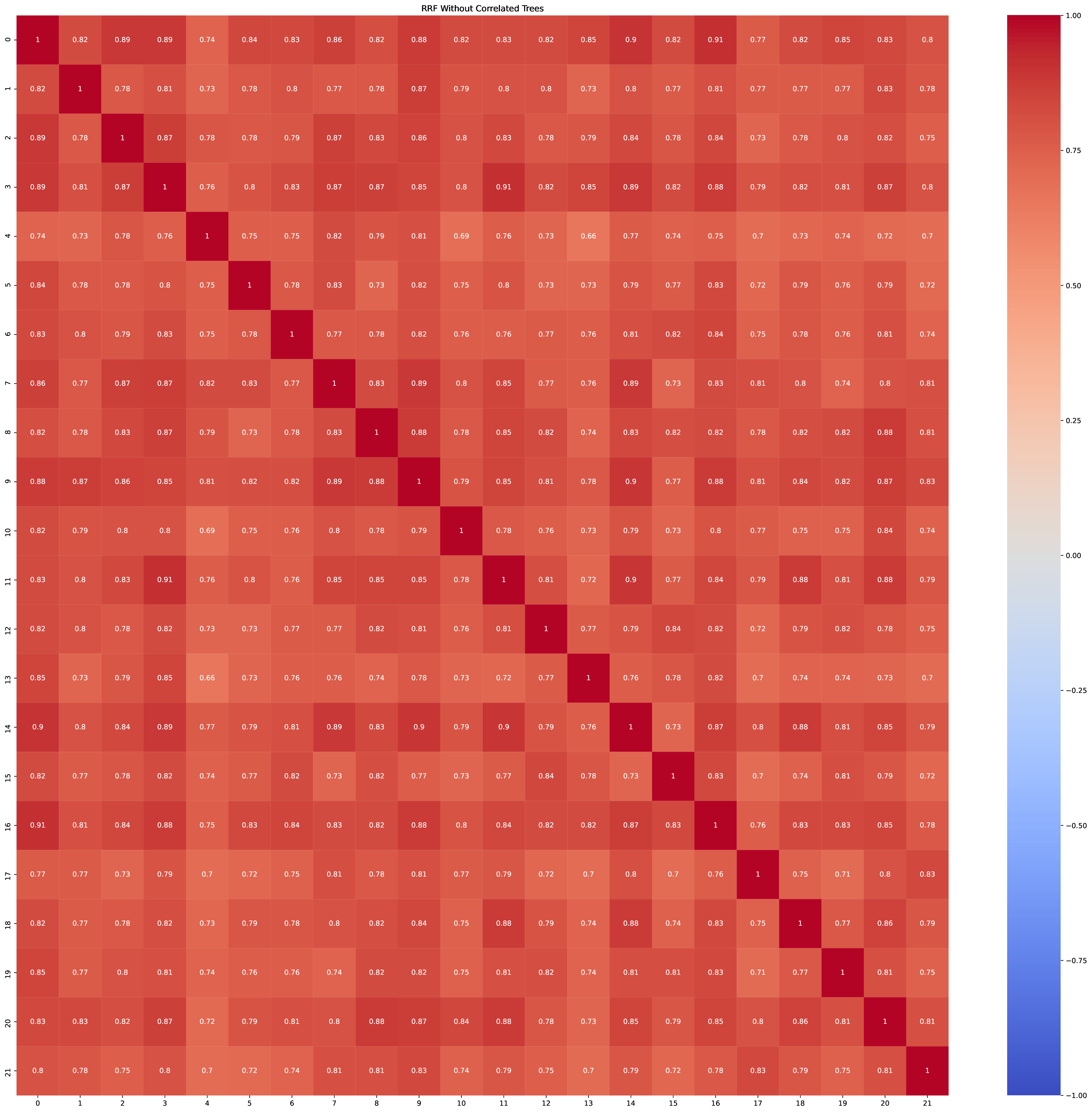}
    \caption{Pairwise correlation heatmaps of tree predictions on the Breast Cancer dataset: Before pruning (left) vs After pruning (right)}
    \label{fig2}
\end{figure}

Figure~\ref{fig2} visualizes how pruning reduces redundancy among trees. In the left heatmap (before pruning), most off-diagonal entries are bright red, showing that many trees make very similar predictions. After grouping highly correlated trees and keeping only the best performer from each group, the right heatmap contains mostly cooler colors, indicating that the remaining trees are substantially less correlated. This increased diversity helps the ensemble generalize better.

Formally, we compute the Pearson correlation between each pair of prediction vectors \(\hat{p}_i, \hat{p}_j\):
\[
r_{i,j}
=
\frac{\sum_{k=1}^{n} (\hat{p}_{i,k}-\bar{\hat{p}}_i)\,(\hat{p}_{j,k}-\bar{\hat{p}}_j)}
     {\sqrt{\sum_{k=1}^{n}(\hat{p}_{i,k}-\bar{\hat{p}}_i)^2}\;\sqrt{\sum_{k=1}^{n}(\hat{p}_{j,k}-\bar{\hat{p}}_j)^2}}.
\]
Trees with \(r_{i,j} > \delta\) are clustered together; within each cluster, only the tree with highest AUC is kept, and the rest are pruned. This process yields a refined ensemble of \(Q\) uncorrelated, high-performing trees.

\section{Results and Discussion} \label{sec3}
This study aimed to enhance the performance of the Random Forest algorithm by designing a self-adjusting Random Forest framework, which selects uncorrelated, high-performing trees using AUC and correlation based clustering. Our research was conducted on four binary and four multiclass datasets. 
% For the Titanic dataset, a total of \textbf{-----} trees were initially generated by the dynamically adjusted model. After selecting only the high-performing and uncorrelated trees, the final model retained \textbf{-----} trees.

We conducted a comprehensive evaluation of the proposed Refined Random Forest (RRF) in comparison to the standard Random Forest (RF). The performance was assessed using eight publicly available datasets from OpenML \textit{Titanic}, \textit{Breast Cancer}, \textit{Diabetes}, \textit{Adult Income}, \textit{Letter}, \textit{OptDigits}, \textit{Sat Image}, and \textit{MNIST} employing the Area Under the Receiver Operating Characteristic Curve (AUC-ROC) as the primary evaluation metric.

Each model was trained and evaluated over multiple randomized iterations per dataset. The number of decision trees varied across different runs. The AUC-ROC metric was chosen due to its robustness and suitability for handling imbalanced datasets, providing a reliable measure of model discrimination performance.

\begin{figure}[h]
    \centering
    \includegraphics[width=1\textwidth]{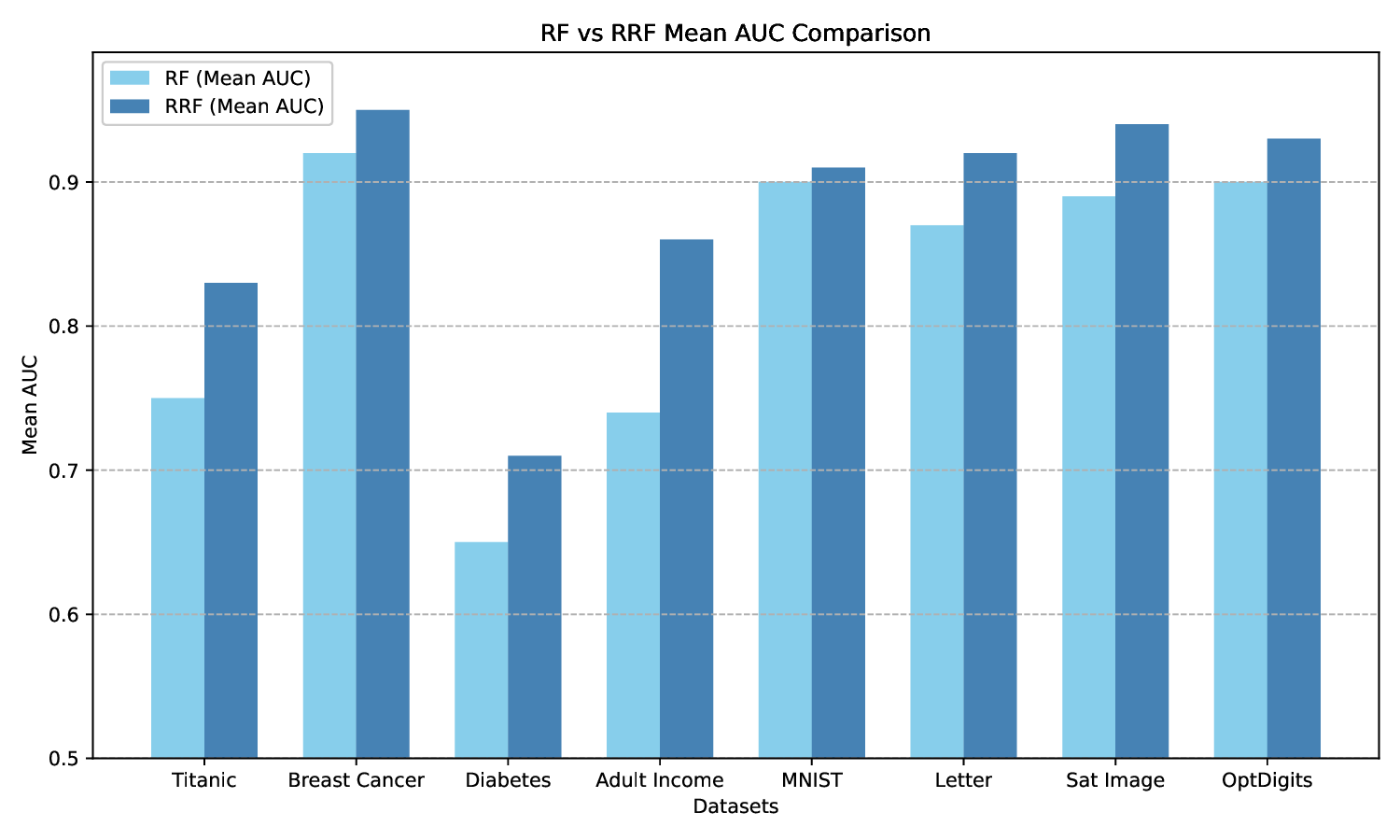}
    \caption{AUC-ROC comparison between Refined Random Forest (RRF) and standard Random Forest (RF) across eight datasets}
    \label{fig:auc_comparison}
\end{figure}
\FloatBarrier

\subsection{Mean Accuracy of RF vs RRF acorss 8 datasets}
To summarize overall performance, we computed the mean accuracy for RRF across each dataset. Table~\ref{tab:mean_auc} reports the mean accuracy of the baseline Random Forest (RF) and our Refined Random Forest (RRF) across all eight datasets. As shown, RRF consistently achieves a higher mean accuracy on every dataset, demonstrating its improved robustness and discriminatory power. Next, we investigated how many trees remain after removing those deemed “correlated” by our pairwise‐correlation threshold test.

\begin{table}[htbp]
    \centering
    \caption{Mean AUC–ROC per dataset}
    \label{tab:mean_auc}
    \begin{tabular}{p{3cm}p{3cm}p{3cm}}
    \toprule
    \textbf{Dataset}       & \textbf{RF (Mean Accuracy)} & \textbf{RRF (Mean Accuracy)} \\ 
    \midrule
    Titanic                & 0.75                   & 0.83                    \\
    Breast Cancer          & 0.92                   & 0.95                    \\
    Diabetes               & 0.65                   & 0.69                    \\
    Adult Income           & 0.74                   & 0.84                    \\
    MNIST                  & 0.90                   & 0.91                    \\
    Letter                 & 0.87                   & 0.91                    \\
    Sat Image              & 0.89                   & 0.94                    \\
    OptDigits              & 0.90                   & 0.93                    \\
    \bottomrule
    \end{tabular}
\end{table}

\begin{table}[htbp]
    \centering
    \caption{Number of trees before and after removing correlated trees for binary datasets}
    \label{tab:binary_trees}
    \begin{tabular}{p{3cm}p{3cm}p{3cm}}
    \toprule
    \textbf{Dataset}      & \textbf{Trees in IRF} & \textbf{Trees in RRF} \\ 
    \midrule
    Breast Cancer         & 44                    & 36                         \\
    Titanic               & 89                   & 55                          \\
    Diabetes              & 172                    & 165                          \\
    Adult Income          & 184                    & 96                         \\
    \bottomrule
    \end{tabular}
\end{table}

\begin{table}[htbp]
    \centering
    \caption{Number of trees before and after removing correlated trees for multiclass datasets}
    \label{tab:multiclass_trees}
    \begin{tabular}{p{3cm}p{3cm}p{3cm}}
    \toprule
    \textbf{Dataset}      & \textbf{Trees in IRF} & \textbf{Trees in RRF}  \\ 
    \midrule
    MNIST                 & 40                        & 35                         \\
    Letter                & 164                        & 152                          \\
    Sat Image             & 40                        & 28                          \\
    OptDigits             & 59                        & 57                          \\
    \bottomrule
    \end{tabular}
\end{table}

Table \ref{tab:binary_trees} lists, for each binary‐class dataset, the total number of trees built by IRF (before pruning) and the number of uncorrelated trees retained after pruning. We see that a significant fraction of trees are pruned away in each case, which helps improve ensemble diversity.

Smilarly, Table \ref{tab:multiclass_trees} shows the analogous counts for the four multiclass datasets. Again, “IRF” denotes the original number of trees grown (before pruning), and “Uncorrelated Trees” is the count after eliminating correlated ones. By comparing Tables \ref{tab:binary_trees} and \ref{tab:multiclass_trees}, one can observe that multiclass datasets often require more aggressive pruning to maintain decorrelated ensembles.

As shown in Table \ref{tab:mean_auc}, RRF consistently outperforms RF on every dataset, with the largest gains on Adult Income (+0.10) and Titanic (+0.08). For image-based multiclass data, Sat Image improves from 0.89 to 0.94 and OptDigits from 0.90 to 0.93.

Tables \ref{tab:binary_trees} and \ref{tab:multiclass_trees} report the number of IRF trees before and after removing correlated ones. On binary datasets, 4 \%–48.9 \% of trees are pruned, while on multiclass datasets about 3.5 \%–30 \% are removed—Sat Image and OptDigits require slightly more pruning due to higher intra-class variability

Finally, the accuracy of the traditional Random Forest was compared with the Refined Random Forest on the final set of trees; our model demonstrates superior accuracy.

\section{Conclusion}
This work set out to investigate whether Random Forest (RF) models can be made more efficient and accurate by dynamically refining both feature sets and tree ensembles. We proposed the Diversity‑Conscious Improved Random Forest (DCIRF), which enhances the classical RF framework by combining iterative feature selection, controlled tree growth, and correlation‑based pruning of redundant trees.

Our major findings demonstrate that DCIRF consistently improves classification accuracy by 3\% to 4\% on eight diverse benchmark datasets including both binary and multiclass tasks. These improvements were achieved without increasing the number of trees used in the final ensemble, confirming that intelligently selecting features and encouraging tree diversity leads to better generalization and resource efficiency.

The relevance and added value of our work lie in its ability to produce a compact yet powerful forest by preserving only those trees that contribute unique, non‑redundant information. This is particularly valuable for real‑world applications where memory, latency, and interpretability are critical.

However, our study is not without limitations. Additionally, the correlation threshold for pruning trees is static and may require tuning per dataset. In addition, the model does not always beat standard random forest on all the available datasets.

Future work will explore adaptive strategies for setting the correlation threshold, extend the method to streaming or online learning scenarios, and examine integration with other ensemble or deep learning techniques. We also recommend applying DCIRF domain specific real world tasks such as medical diagnostics, financial risk analysis, and environmental monitoring to further validate its utility.

\section{Statements \& Declarations} \label{sec4}

\subsection{Funding}
No funding was received to assist with the preparation of this manuscript.

\subsection{Author Contributions}

Saurav Bhandari and Sijan Bhattarai created and built the main DCIRF framework, which included developing the dynamic Random Forest with feature ranking, updating feature weights using the information gain ratio, adding trees in steps, and writing the manuscript. Girija Bhusal implemented correlation-based pruning (cluster formation and selection of optimized trees) and conducted final testing on multiple benchmark datasets. Professor Saroj Shakya provided continuous guidance and supervision throughout the research process, offering valuable insights during the design, implementation, and refinement of the proposed method. Tapendra Pandey helped with manuscript writing and algorithm evaluation.
All authors reviewed and approved the final manuscript.

\subsection{Data availability}
The code and data supporting the findings of this study are available at \url{https://github.com/sauravb777/DCRRF.git}.  
All scripts used to preprocess the datasets, train and evaluate the models, and generate the figures are included in the repository.  

% \bibliography{sn-bibliography}% common bib file


\end{document}